# Towards Mining Effective Pedagogical Strategies from Learner–LLM Educational Dialogues

Liqun He[1[0000-0002-0837-5857]], Manolis Mavrikis[1[0000-0003-0575-0823]], Mutlu Cukurova[1[0000-0001-5843-4854]]

[1]University College London, London, United Kingdom
{liqun.he,m.mavrikis,m.cukurova}@ucl.ac.uk

**Abstract.** Dialogue plays a crucial role in educational settings, yet existing evaluation methods for educational applications of large language models (LLMs) primarily focus on technical performance or learning outcomes, often neglecting attention to learner–LLM interactions. To narrow this gap, this AIED Doctoral Consortium paper presents an *ongoing* study employing a dialogue analysis approach to identify effective pedagogical strategies from learner–LLM dialogues. The proposed approach involves dialogue data collection, dialogue act (DA) annotation, DA pattern mining, and predictive model building. Early insights are outlined as an initial step toward future research. The work underscores the need to evaluate LLM-based educational applications by focusing on dialogue dynamics and pedagogical strategies.

## 1    Problem Statement

With the growing use of large language models (LLMs) in various educational applications, evaluating their quality has become an increasingly important task. Existing studies primarily focus on the technical performance or learning outcome-related measures. Specifically, in the field of computer science, *technical evaluations* focus on LLMs' performance predominantly using non-interactive benchmarks (e.g., accuracy of single-turn outputs) [13]. In the field of education, researchers focus on *learning outcomes* following their use, including affective factors such as anxiety [23], willingness [5], and self-confidence [2], as well as cognitive aspects like knowledge and skill development [4, 21].

However, existing evaluations often overlook the dialogue processes between learners and LLMs, which is both necessary and important to consider. On the one hand, recent studies have shown that superior single-turn performance does not necessarily lead to better multi-turn interactions [11, 13], highlighting the need to move beyond single-turn evaluations. On the other hand, prior research on human tutoring dialogues has demonstrated that patterns of dialogue acts are correlated with student learning gains [3, 25], making dialogue analysis a promising approach for understanding and evaluating the pedagogical effectiveness of LLMs. Through such analysis, we can gain a deeper understanding of the *mechanisms* through which educational LLMs contribute to learning gains. For instance, identifying when and which dialogue acts trigger student motivation or self-reflection may inform the design of educational LLMs that promote deeper engagement and learning.

Towards narrowing the gap, this study proposes a method for mining effective pedagogical strategies from learner–LLM dialogues. Using data from English language learners interacting with a voice chatbot for speaking practice, it investigates the following research question: **'What are the sequential learner-LLM dialogue patterns in effective and less effective tutoring sessions?'**

Through this preliminary work, we seek to deepen the understanding of learner-LLM dialogue dynamics, identify effective pedagogical strategies, and potentially contribute to the development of educational LLM evaluation methods by emphasising dialogue dynamics and pedagogical strategies as focal points.

## 2      Theoretical Basis: Content, Intention, and Speech Act Theory

Searle's Speech Act Theory serves as the foundational theoretical framework for dialogue analysis in this study, outlining three key principles:

**(1) An utterance conveys both its content and intention.** Firstly, Speech Act Theory posits that when we speak, we are not just conveying meanings but also aiming to achieve specific goals [19, 20]. That is, every utterance has a *dual* function: it conveys both a semantic representation (the content) and a pragmatic identification (the intention) [7].

**(2) The content of an utterance is the result of its hidden intention.** Secondly, the theory clarifies the relationship between content and intention: the literally expressed content is the result of the speaker's hidden intentions. An utterance, therefore, is more than just a collection of words; it represents an action carrying meanings beyond its literal interpretation, reflecting the speaker's underlying intentions [19]. For example, when a friend says, 'I am hungry', he or she is not just describing their physical state but aiming to communicate their desire to have lunch.

**(3)  The hidden intention can be inferred from the expressed content.** Third, the theory emphasises that understanding intention is more important than content alone, and suggests that the hidden intention can be inferred from the expressed content. In this theoretical framework, the intention behind an utterance is referred to as a 'speech act' (or interchangeably 'dialogue act' [22], 'dialogue move' [24], or 'communicative act' [9]; in this study, we use the term 'dialogue act'). A dialogue act (DA) represents the 'function of the meaning of the sentence' [19].  further explain that to fully comprehend an utterance, one must go *beyond* its surface-level meaning and uncover the intention behind it (i.e., to identify its DA).

According to Speech Act Theory, tutors' intentions (or, in the context of this paper, the intentions of educational LLMs) are conveyed and reflected by what they say and when they say it [1, 7] and through various forms of dialogue, tutors (or educational LLMs) achieve various pedagogical goals [17]. This framework provides a theoretical foundation for inferring pedagogical intentions by analysing dialogue turns generated by LLMs or expressed by learners, enabling the analysis of pedagogical strategies from learner-LLM dialogues.

## 3     Related Work

**Learning Analytics Approach.** Prior to the widespread adoption of educational LLM applications, researchers had already analysed human-human DAs to evaluate and predict tutoring quality. For example, [14] examined 50 tutoring sessions across mathematics, chemistry, and physics, applying a BERT model for DA annotation and the TraMineR package for sequential pattern mining. Their findings showed successful sessions—those resulting in correct or partially correct student solutions—featured tutors using more thought-provoking questions and informational hints, and fewer irrelevant statements. Similarly, [3] analysed 2,250 minutes of one-to-one online mathematics tutoring using manual annotations, primarily focusing on tutors' dialogic behaviours but also including other interactive behaviours (e.g., writing on the virtual classroom interface). Sequential pattern mining (CM-SPAM algorithm) identified effective pedagogical strategies, such as prompting student self-correction and allowing appropriate pauses for reflection. These patterns then successfully informed decision-tree classifiers (JRIP and J-48) to distinguish tutor effectiveness. This work demonstrates the potential of behavioural action patterns for both identifying effective pedagogical strategies and monitoring the quality of online tutoring sessions.

**LLM Evaluation for Dialogue Tasks.** Within computer science, recent studies have also begun moving beyond automatic metrics alone, incorporating human evaluations to measure LLM performance on various dialogue tasks. For example, [13] conducted *turn-level* human surveys evaluating emotional connections in social dialogues (e.g., empathy, fluency, sensibility, and specificity). Similarly, Google's LearnLM team [10] assessed the pedagogical effectiveness of educational LLMs in conversational tutoring through learner and pedagogical rater feedback at both *turn-level* (e.g., "Does this turn inspire interest?") and *conversation-level* (e.g., "Which tutor more closely resembles an excellent human tutor?"). Follow-up research [12] further compared LLM models' pedagogical performance by inviting experienced tutors to complete *conversation-level* questionnaires after reviewing learner-LLM conversation transcripts.

**Limitations of Previous Research.** Previous analyses of human-human dialogues offered valuable methodological insights for assessing human-LLM interactions. However, [18] found significant disparities between AI-generated and human-authored texts. Given these differences and the shift from human-human to human-LLM learning environments, previous findings cannot be directly applied and require further validation.

On the other hand, recent developments in LLM evaluation, despite beginning to focus on dialogue, still fail to effectively capture pedagogical strategies. A key reason is that pedagogical strategies are typically represented by *sequences* of dialogue acts rather than isolated dialogue turns [15]. For example, the 'scaffolding' strategy—where learners work independently while tutors provide minimal assistance—can only be identified through a series of hints delivered in the form of questions. Current LLM evaluations primarily operate at either the *micro* (single-turn) or *macro* (full-conversation) levels. However, micro-level evaluations may be too narrow to capture pedagogical strategies, while macro-level evaluations may be too broad to assess LLMs' ability to implement specific strategies and follow pedagogical instructions. Focusing on DA patterns offers a potential middle ground (a *meso*-level) between atomistic and holistic approaches, providing a more balanced dimension for understanding and evaluating pedagogical strategies in learner-LLM dialogues.

## 4       Proposed Solution: A Dialogue Approach

The proposed solution aims at employing a dialogue analysis approach to identify predominant pedagogical strategies in learner-LLM interactions, consisting of four key components:

**(1) Dialogue Data Collection.** Establishing an accessible, secure, and user-friendly method for storing and managing dialogue data across multiple students and sessions is the first step in enabling further analysis. To achieve this, we developed a platform that interfaces with LLMs via application programming interfaces (APIs) and securely stores dialogue data in a cloud database. Each recorded dialogue entry includes a unique identifier, timestamp, participant role (student or chatbot), model name, user ID, session ID (defined as continuous dialogue with gaps under 15 minutes), and dialogue transcript content.

**(2) Dialogue Act Annotation.** Language is a complex system, making it difficult to infer pedagogical intentions (i.e., DAs). Coding systematically addresses this by developing a coding scheme with descriptors (codes) to capture key characteristics of utterances and applying it to annotate dialogue turns [9, 26], thereby converting transcribed dialogues into structured and analysable data [15]. For this purpose, we adapted a preliminary coding scheme tailored to our context of oral English practice. This scheme is primarily based on the SETT coding scheme [27, 28] and the Cambridge Educational Dialogue Research Group's SEDA [9] and T-SEDA [26] coding schemes. We plan to further incorporate aspects from the latest Tech-SEDA for technology-mediated contexts [8]. Given that this is the pilot phase, we selected only three students with the most noticeable progress (High Progress (HP) group) and three students with the least progress (Low Progress (LP) group). We then collected dialogue data from each student's practice sessions on the first and last day, resulting in a total of 6 dialogues per group. The average number of dialogue turns per session was similar between the HP group (83.3 turns) and the LP group (86.7 turns). However, students in the HP group produced longer utterances, averaging 15.3 words per turn compared to 11.5 words in the LP group. In total, we manually annotated 1,020 dialogue turns across both groups for subsequent pattern mining.

**(3) Dialogue Act Pattern Mining.** To identify predominant DA patterns, we then applied the CM-SPAM algorithm [6] with a minimum pattern length of 2, a maximum gap of 1, and a minimum support of 50%. An example of insights gained from the results is as follows: the HP group exhibited the pattern [t]Q - [s]R - [s]Q (bot

question–student response–student question), while the LP group uniquely showed [t]Q–[s]Q. Transcript analysis further suggests that the absence of [s]R in the LP group indicates learners' difficulty in understanding the initial question ([t]Q), prompting students to ask a new question rather than respond.

**(4) Predictive Model Building.** Building upon identified dialogue patterns as indicators of pedagogical strategies, our next step involves training predictive models (e.g., decision tree classifiers following [3], or Bayesian Networks as described in [16]). Such models provide interpretable insights, enabling systematic assessment and targeted improvement of educational LLM systems.

## 5    Expected Contributions

**(1) For LLM System Design.** This proposed method seeks to identify predominant DA patterns in learner-LLM interactions, providing insights into both effective and ineffective pedagogical strategies within LLM-mediated learning environments. These insights can potentially support educators, designers, and researchers in understanding learner-LLM interaction dynamics, informing and optimising instructional design.

**(2) For LLM System Evaluation.** This ongoing work highlights the need for evaluating educational LLMs by focusing on learner-LLM dialogue dynamics, moving beyond solely single-turn evaluations or learning gains. This potentially contributes to the advancement of methods for evaluating the pedagogical performance of LLMs.